\documentclass{article}
\usepackage[nonatbib,dblblindworkshop,final]{neurips_2025}
\usepackage[T1]{fontenc}
\usepackage{multirow}
\usepackage{booktabs}
\usepackage{microtype}
\usepackage{amsmath}
\usepackage{bbm}
\usepackage{amssymb}
\usepackage{graphicx}
\usepackage[table]{xcolor}
\usepackage{hyperref,url}
\usepackage{xcolor}
\usepackage{blindtext}

\usepackage[backend=bibtex,sorting=none,maxbibnames=20]{biblatex}
\begin{document}

\title{Nonlinear Optimization with GPU-Accelerated Neural Network Constraints}
\workshoptitle{ScaleOpt: GPU-Accelerated and Scalable Optimization}
% Alternate titles:
% - Nonlinear optimization with GPU-accelerated NN constraints
% - Nonlinear optimization with GPU-accelerated embedded NNs
% - Nonlinear optimization with embedded NNs on GPUs
% - Nonlinear optimization with (embedded) NN evaluation on GPUs

\author{
  Robert Parker      \\ %\orcidID{0000-1111-2222-3333}
  Los Alamos National Laboratory \\
  Los Alamos, NM 87545, USA \\
  \And Oscar Dowson     \\ %\orcidID{1111-2222-3333-4444}
  Dowson Farms \\
  New Zealand \\
  \And Nicole LoGiudice \\ %\orcidID{2222-3333-4444-5555}
  Texas A\&M University \\
  College Station, TX 77843, USA \\
  \And Manuel Garcia    \\ %\orcidID{2222-3333-4444-5555}
  Los Alamos National Laboratory \\
  Los Alamos, NM 87545, USA \\
  \And Russell Bent     \\ %\orcidID{2222-3333-4444-5555}
  Los Alamos National Laboratory \\
  Los Alamos, NM 87545, USA \\
}
%
%\institute{
%  Los Alamos National Laboratory, Los Alamos, NM 87544, USA
%  %\email{\{rbparker,mjgarcia,rbent\}@lanl.gov}\\
%  \and Dowson Farms\\
%  %\email{oscar@dowsonfarms.co.nz}\\
%  \and Texas A\&M University, College Station, TX 77843, USA\\
%  %\email{nicolelogiudice30@tamu.edu}
%}
%
\maketitle              % typeset the header of the contribution
\begin{abstract}
We propose a reduced-space formulation for optimizing over trained
neural networks where the network's outputs and derivatives are
evaluated on a GPU. To do this, we treat the neural network
as a ``gray box'' where intermediate variables and constraints
are not exposed to the optimization solver.
Compared to the full-space formulation, in which 
intermediate variables and constraints \emph{are} exposed to the
optimization solver, the reduced-space formulation leads to
faster solves and fewer iterations in an interior point method.
We demonstrate the benefits of this method on two optimization problems:
Adversarial generation for a classifier trained on MNIST images
and security-constrained optimal power flow with transient feasibility
enforced using a neural network surrogate.
\end{abstract}

\section{Introduction}
Optimization over trained machine learning (ML) models can be used to verify
ML models \cite{tjeng2018evaluating,bunel2018unified},
generate adversarial examples, and use these ML models for
optimization-based design and control \cite{bugosen2024,lopezflores2024}.
For smooth, nonlinear ML models (e.g., neural networks with non-ReLU activation
functions) or design or control tasks involving nonconvex constraints,
the resulting optimization problem must be solved with a nonconvex optimization
solver (as opposed to a mixed-integer, linear solver).
Global optimization of nonconvex functions is well-known to be NP-hard,
but local optimization of these functions can be done efficiently with
interior point methods \cite{nocedalwright}.

Despite favorable scalability of interior point methods in other domains
(e.g., power grid operation \cite{aravena2023}),
we are unaware of recent work investigating the scalability of interior point methods
with large neural networks (NNs) embedded. Recent work using
interior point methods to optimize over trained NNs
\cite{kilwein2023,bugosen2024,casas2025comparison}
has been limited to small NN models (fewer than 1M trained parameters).
In this work, we examine the scalability of interior point methods with large
NN models (over 100M trained parameters) embedded.
We propose a reduced-space formulation with GPU acceleration as a scalable
method of solving these optimization problems.

GPU-accelerated optimization has received much recent attention.
First-order methods for linear programming \cite{applegate2021pdhg}
have led to implementations that offload matrix-vector multiplications
to a GPU \cite{lu2024cupdlp}. Simultaneously, the development of
GPU-accelerated sparse matrix factorization algorithms, e.g., cuDSS \cite{cudss},
have enabled GPU-acceleration for a wider class of optimization problem
\cite{shin2024accelerating,pacaud2024gpu}.
In contrast to these general-purpose methods, our method is tailored
to nonlinear optimization problems with neural networks embedded.
%In contrast to these general-purpose methods, our method is tailored to
%optimization problems with neural networks embedded.
%Instead of developing a new solver to exploit the neural network structure,
%we exploit this structure only in function and derivative evaluations,
%allowing us to interface with existing nonlinear optimization solvers
%while still gaining the benefits of GPU acceleration.

\section{Nonlinear optimization}\label{sec:nonlinear_opt}
We study nonlinear optimization problems in the form given by
Problem \eqref{eqn:nlopt}:
\begin{equation}
  \min_x f(x) \text{ subject to } \left\{
    \begin{array}{l}g(x) = 0\\ x\geq 0\end{array}\right.
  %\begin{array}{cl}
  %  \displaystyle \min_{x} & f(x) \\
  %  \text{subject to} & g(x) = 0\\
  %  & x \geq 0 \\
  %\end{array}
  \label{eqn:nlopt}
\end{equation}
%We consider lower bounds of zero for simplicity, but general upper and lower
%bounds can be handled easily.
%Problems with general inequality constraints are often reformulated into this form
%by the addition of slack variables, which we assume has already happened.
%
%General purpose open-source software is available for the solution of \eqref{eqn:nlopt}.
%In this paper, we use IPOPT \cite{ipopt}.
%As input, IPOPT requires oracles for
%the objective function $f(x): {\mathbb{R}}^N \rightarrow \mathbb{R}$, the gradient
%of the objective $\nabla f(x): {\mathbb{R}}^N \rightarrow \mathbb{R}^N$, the constraints
%$g(x): {\mathbb{R}}^N \rightarrow \mathbb{R}^M$, the Jacobian of the constraints
%$\nabla g(x): {\mathbb{R}}^N \rightarrow \mathbb{R}^{M\times N}$, and the \textit{Hessian-of-the-Lagrangian}
%$\nabla^2\mathcal{L}(x): {\mathbb{R}}^N \rightarrow \mathbb{R}^{N\times N}$, which is defined as:
%$$\nabla^2\mathcal{L}(x) = \sigma \nabla^2 f(x) + \sum\limits_{i=1}^M \lambda_i \nabla^2 g_i(x),$$
%where $\lambda$ is the Lagrangian multiplier associated with the equality constraint,
%and $\sigma$ is a scalar constant. Computing the first- and second-order derivatives
%of $f$ and $g$, and evaluating all the oracles is a potential bottleneck in the solution process.
%
This formulation does not preclude general inequality constrains, which may be reformulated
with a slack variable.
We consider interior point methods, such as IPOPT \cite{ipopt}, for solving \eqref{eqn:nlopt},
which require that functions $f$ and $g$ are twice continuously differentiable \cite{nocedalwright}.
We note that this requirement precludes neural networks with non-differentiable activation
functions such as ReLU.
Interior point methods iteratively compute search directions $d$ by solving the linear system
\eqref{eqn:linear_system}:
%With a slight abuse of notation to simplify unnecessary details, IPOPT's solution method
%for \eqref{eqn:nlopt} is an iterative algorithm that, at each iteration, involves the
%solution of the following linear system, which is related to the KKT conditions of \eqref{eqn:nlopt}:
\begin{equation}
  \begin{bmatrix}
    (\nabla^2\mathcal{L}(x)+\alpha) & \nabla g(x)^T \\
    \nabla g(x) & 0
  \end{bmatrix} d = -\begin{bmatrix}
      \nabla f(x) + \nabla g(x)^T \lambda + \beta\\
      g(x)
  \end{bmatrix},
  \label{eqn:linear_system}
\end{equation}
where $\alpha$ and $\beta$ are additional terms that are not shown for simplicity. The matrix on the left-hand side is referred to as the
{\it Karush-Kuhn Tucker}, or {\it KKT}, matrix.
%Solving this system, which
%is achieved through a factorization using third-party subroutines, is another potential
%bottleneck in the solution process.
To construct this system, solvers rely on callbacks (oracles) that provide the Jacobian, $\nabla g$,
and the Hessian of the Lagrangian function, $\nabla^2\mathcal{L}$. These are typically
provided by an automatic differentiation system \cite{griewank2008}.

\section{Representing neural networks as constraints}\label{sec:representing}
We consider neural network predictors, $y={\rm NN}(x)$,
defined by repeated application of an affine transformation and a nonlinear
activation function $\sigma$ over $L$ layers:
\begin{equation}
  \begin{aligned}
    y_l &= \sigma_l (W_l y_{l-1} + b_l) && l \in \{1,\dots,L\},
    \label{eqn:nn}
  \end{aligned}
\end{equation}
where $y_0 = x$ and $y = y_L$.
In this context, training weights $W_l$ and $b_l$ are considered constant.
Pre-trained neural network predictors may be embedded into the constraints
of an optimization problem using either \textit{full-space} or \textit{reduced-space} formulations
\cite{omlt,schweidtmann2019deterministic}.

\subsection{Full-space}

%In the \textit{full-space} formulation, we add an intermediate vector-valued decision
%variable $z_l$ to represent the output of the affine transformation in each layer
%$l$, and we add a vector-valued decision variable $y_l$ to represent the output of
%each nonlinear activation function. We then add a linear equality constraint
%to enforce the relationship between $y_{l-1}$ and $z_l$ and a
%nonlinear equality constraint to enforce the relationship between $z_l$ and $y_l$.
%Thus, the neural network in \eqref{eqn:nn} is encoded by the constraints:
In the \textit{full-space} formulation, we add additional variables and constraints
for each layer of the neural network:
\begin{equation}
  \begin{aligned}
    z_l &= {W_l} y_{l-1} + b_l && l \in \{1,\dots,L\} \\
    y_l &= \sigma_l(z_l) && l \in \{1,\dots,L\}.
    \label{eqn:nn_full}
  \end{aligned}
\end{equation}
The full-space approach prioritizes small, sparse nonlinear constraints
at the cost of introducing (potentially) many new variables and constraints.

\subsection{Reduced-space}

In the \textit{reduced-space} formulation, we add a single vector-valued decision variable
$y$ to represent the output of the final activation function and a single vector-valued
nonlinear equality constraint that encodes the complete neural network:
\begin{equation}
    y = {\rm NN}(x) = \sigma_L({W_L} (\ldots\sigma_l({W_l} (\ldots\sigma_1({W}_1 x + b_1) \ldots) + b_l)\ldots) + b_L)
    \label{eqn:nn_reduced}
\end{equation}
The benefit of this formulation is that we only add a single decision variable and a single
nonlinear equality constraint (each of dimension of the neural network's output).
The drawback is that this nonlinear equality constraint contains a large, complicated
algebraic expression that can be expensive to evaluate and differentiate.
However, this can be alleviated by using dedicated neural network modeling software
(e.g., PyTorch \cite{paszke2019pytorch}) to represent the neural network constraint
rather than general-purpose algebraic modeling software (e.g., JuMP \cite{jump1}).
Doing so also allows us to exploit GPU acceleration via PyTorch's CUDA interface.
We note that this approach limits the information that can be communicated between
the neural network and the optimization solver. Oracles (which PyTorch provides
for function, Jacobian, and Hessian evaluation) can be queried, but the internal
structure of the neural network cannot be exploited directly.
This is suitable for nonlinear \textit{local} optimization, where only oracles
are required, but not for \textit{global} optimization, where the algebraic
form of constraints is exploited to construct relaxations.

% TODO
\subsubsection{The Hessian of the Lagrangian}
Interior point methods require oracles to evaluate the Hessian of the
Lagrangian, $\nabla^2\mathcal{L}$:
\begin{equation}
  \nabla^2\mathcal{L}(x,\lambda) = \nabla^2 f(x) + \sum_i \lambda_i \nabla^2 g_i(x),
  \label{eqn:hess-lag}
\end{equation}
where $\lambda$ is the vector of Lagrange multipliers of the equality constraints
in Equation \ref{eqn:nlopt}.
Naively constructing $\sum_i\lambda_i \nabla^2 g_i$ for constraints involving
a reduced-space neural network, $g(x,y)=y-{\rm NN}(x)$, would require
(1) evaluating the Hessian of the neural network, $\nabla^2{\rm NN}(x)$,
and (2) computing a sum-product with $\lambda$ along the first rank
of this third-order Hessian tensor. This sum-product is potentially
expensive for a dense Hessian ($\mathcal{O}(mn^2)$, where $m$ is the output
dimension and $n$ is the input dimension of the neural network).
For this reason, we encode the Lagrangian of the neural network,
$\lambda^T {\rm NN}(x)$, directly as a linear layer in PyTorch
and differentiate through this scalar-valued function to directly
compute the $n\times n$ Hessian matrix, $\sum_i \lambda_i \nabla^2 {\rm NN}_i(x)$.

%We evaluate the Hessian of the Lagrangian directly rather than constructing
%it from all individual constraint Hessians because the former approach
%is more efficient (see Supplemental Information).

\section{Test problems}
We test the full and reduced-space formulations on two nonlinear optimization
problems with NNs embedded: (1) Adversarial image generation using a neural
network MNIST classifier (denoted ``MNIST'') and (2) security-constrained
optimal power flow with a neural network constraint enforcing transient
feasibility (denoted ``SCOPF'').

\begin{table}
  \rowcolors{1}{gray!10}{white}
  \centering
  \caption{Structure of neural network models}
  \begin{tabular}{cccccc}
    \toprule
Model & N. inputs & N. outputs & N. neurons & N. param. &      Activation \\
\midrule
MNIST &       784 &         10 &         1k &      167k &    Tanh+SoftMax \\
MNIST &       784 &         10 &         3k &        1M &    Tanh+SoftMax \\
MNIST &       784 &         10 &         5k &        5M &    Tanh+SoftMax \\
MNIST &       784 &         10 &        11k &       18M &    Tanh+SoftMax \\
MNIST &       784 &         10 &        21k &       70M &    Tanh+SoftMax \\
MNIST &       784 &         10 &        41k &      274M & Sigmoid+SoftMax \\
\midrule
SCOPF &       117 &         37 &        254 &       15k &            Tanh \\
SCOPF &       117 &         37 &         1k &      578k &            Tanh \\
SCOPF &       117 &         37 &         5k &        4M &            Tanh \\
SCOPF &       117 &         37 &        36k &       68M &            Tanh \\
SCOPF &       117 &         37 &       152k &      592M &            Tanh \\
\bottomrule
  \end{tabular}
  \label{tab:nns}
\end{table}

\subsection{Adversarial image generation for an MNIST classifier}
We train a set of neural networks using smooth activation functions
(hyperbolic tangent, sigmoid, and softmax) to serve as classifiers for images from
the MNIST set of handwritten digits \cite{lecun1998mnist}.
Inputs are the 28$\times$28 grayscale pixel colors, flattened into a 784-dimensional
vector, and outputs are scores for each digit, 0-9, that may be interpreted as the
probability that the image represents the corresponding digit. The neural networks
each have seven layers total and have between 128 and 8192 neurons per hidden layer.
The networks are trained to have accuracies of at least 95\% on the test set
of 10,000 images from the dataset.
The number of neurons, trained parameters, and activation functions for each
network are shown in Table \ref{tab:nns}.
Our optimization problem finds a minimal perturbation to a reference image
that results in a misclassification:
\begin{equation}
  \begin{array}{cll}
    \displaystyle\min_x & \left\| x - x_{\rm ref} \right\|_1 \\
    \text{subject to} & y = {\rm NN}(x) \\
    & y_t \geq 0.6 \\
  \end{array}
  \label{eqn:adversarial-image}
\end{equation}
Here, $x$ contains the grayscale values of the generated image, $x_{\rm ref}$
contains those of the reference image, and $t$ is the coordinate of the neural network's output,
$y$, corresponding to a target label (misclassification). The neural network
constraint, $y={\rm NN}(x)$, may be written in full-space or reduced-space formulations.
The image must be misclassified with at least 60\% confidence.

\subsection{Transient-constrained optimal power flow}
Security-constrained optimal power flow (SCOPF) is a well-established problem
for dispatching generators in an electric power network in which feasibility
of the network (i.e., the ability to meet demand) is enforced for a set
of {\it contingencies} \cite{aravena2023}.
Each contingency $k$ represents the loss of a set of generators and/or power lines.
We consider a variant of this problem where, in addition to enforcing steady-state
feasibility, we enforce feasibility of the transient response resulting from
the contingency. In particular, we enforce that the transient frequency at each
bus is at least ${\bf \eta}=59.4$ Hz for the 30 second interval following each contingency.
This problem is given by Equation \ref{eqn:opf}:
\begin{equation}
  \min_{S^g,V} c(\mathbb{R}(S^g)) \text{ subject to }
  \left\{\begin{array}{ll}
      F_k (S^g, V, {\bf S^d}) \leq 0 & k \in \{0,\dots,K\} \\
      G_k(S^g, {\bf S^d}) \geq \mathbf{\eta}\mathbbm{1}& k\in \{1,\dots,K\}. \\
  \end{array}\right.
  %\begin{array}{cll}
  %  \displaystyle \min_{S^g,V,S} & c(\mathbb{R}(S^g)) \\
  %  %\text{s.t.} & {\bf v^l} \leq \left| V_i \right| \leq {\bf v_i^u} & i\in N\\
  %  %& {\bf S^{gl}_i} \leq S_i^g \leq {\bf S_i^{gu}} & i \in N \\
  %  %& \left|S_{ij}\right| \leq {\bf S_{ij}^u} & (i,j) \in E\\
  %  %& \displaystyle S_i^g - {\bf S_i^d} = \sum_{(i,j)\in E\cup E^R} S_{ij} & \text i\in N \\
  %  %& S_{ij} = {\bf Y^*_{ij}} V_i V_i^* - {\bf Y^*_{ij}} V_i V_j^* & (i,j)\in E\cup E^R \\
  %  %& {\bf \theta^l_{ij}} \leq \angle (V_i V_j^*) \leq {\bf \theta^u_{ij}} & (i,j)\in E \\
  %  \text{s.t.} & F_k (S^g, V, {\bf S^d}) \leq 0 & k\in K \\
  %  & G_k (S^g, {\bf S^d}) \geq {\bf \eta} \mathbbm{1} & k\in K\\
  %\end{array}
  \label{eqn:opf}
\end{equation}
Here, $S^g$ is a vector of complex AC power generations for each generator in the network,
$V$ is a vector of complex bus voltages, $c$ is a quadratic cost function, and
$\bf S^d$ is a constant vector of complex power demands.
Here $F_k \leq 0$ represents the set of constraints enforcing feasibility of the
power network for contingency $k$ (see \cite{cain2012history}), where $k=0$ refers to the base network,
and $G_k$ maps generations and demands to the minimum frequency at each bus over the
interval considered.

In this work, we consider an instance of Problem \ref{eqn:opf} defined on a 37-bus
synthetic test grid \cite{birchfield2017,hawaii40}.
In this case, $G_k$ has 117 inputs and 37 outputs.
We consider a single contingency that outages generator 5 on bus 23. We choose a small
grid model with a single contingency because our goal is to test the different neural
network formulations, not the SCOPF formulation itself.
\\

%\subsubsection{Stability surrogate model}
{\bf Stability surrogate model}~
Instead of considering the differential equations describing transient behavior of the
power network directly in the optimization problem, we approximate $G_k$ with a neural
network surrogate, as proposed by \textcite{garcia2025transient}.
Our surrogate is trained on data from 110 high-fidelity simulations using PowerWorld \cite{powerworldmanual}
with generations and loads uniformly sampled from within a $\pm 20\%$ interval of each nominal value.
We use sequential neural networks with $\tanh$ activation functions with between
two and 20 layers and between 50 and 4,000 neurons per layer.
As shown in Table \ref{tab:nns}, these networks have between 7,000 and 592 million trained parameters.
The networks are trained to minimize mean squared error using the Adam optimizer \cite{adam}
until training loss (mean squared prediction error) is below 0.01 for 1,000 consecutive epochs.
We use a simple training procedure and small amount of data because our goal is
to test optimization formulations with embedded neural networks, rather than the neural
networks themselves.

%Because the focus of this work is to test different formulations for trained neural
%networks in optimization problems, we do not focus extensively on validating the
%neural network model. In particular, we do not validate the behavior of the neural network
%surrogate on out-of-sample generation and load data.

\section{Results}

\subsection{Computational setting}
We model the SCOPF problem using PowerModels \cite{powermodels},
PowerModelsSecurityConstrained \cite{powermodelssecurity}, and JuMP \cite{jump1}.
Neural networks are modeled using PyTorch \cite{paszke2019pytorch} and embedded into
the optimization problem using MathOptAI.jl \cite{dowson2025moai}.
Optimization problems are solved using the IPOPT interior point method \cite{ipopt}
with MA57 \cite{ma57} as the linear solver to a tolerance of $10^{-6}$.
We note that IPOPT runs exclusively on CPU. Interfacing our function evaluation
methods with a GPU-enabled nonlinear optimization solver, e.g., MadNLP \cite{shin2024accelerating},
would be valuable future work.
We evaluate and differentiate the full-space model's functions on a CPU using JuMP.
While we could use a framework that supports GPU-accelerated function evaluations
such as PyTorch or ExaModels (see \cite{shin2024accelerating}), our results indicate that these function evaluations
are not a significant contributor to solve time (see Table \ref{tab:solvetime-breakdown}).
Because our reduced-space models use PyTorch, they can be evaluated on a CPU or GPU.
We run our experiments on the Darwin cluster at Los Alamos National Laboratory.
The CPU used in these experiments is an AMD EPYC with 128 cores and 500 GB
of RAM and the GPU is an NVIDIA A100 with 40 GB of on-device memory.
The code used to produce these results can be found at \url{https://github.com/Robbybp/moai-examples}.

\subsection{Structural results}

Table \ref{tab:structure} shows the numbers of variables, constraints, and
nonzero entries of the derivative matrices for the two optimization problems with
different neural networks and formulations.
With the reduced-space formulations, the structure of the optimization problem does not change
as the neural network surrogate adds more interior layers.
By contrast, the full-space formulation grows in numbers of variables, constraints,
and nonzeros as the neural network gets larger.
In this case, the number of nonzeros in the Jacobian matrix is approximately the number
of trained parameters in the embedded neural network model.
These problem structures suggest that the full-space formulation will lead to expensive
KKT matrix factorizations, while this will not be an issue for the reduced-space formulation.

\begin{table}%[h!]
  \centering
  \rowcolors{1}{gray!10}{white}
  \caption{Structure of optimization problems with embedded neural networks}
  \begin{tabular}{ccccccc}
    \toprule
    Model &   Formulation &        NN param. & N. var. & N. con. & Jac. NNZ & Hess. NNZ \\
\midrule
MNIST &    Full-space &      167k &      3k &      2k &     171k &       661 \\
MNIST &    Full-space &        1M &      7k &      5k &       1M &        2k \\
MNIST &    Full-space &        5M &     12k &     11k &       5M &        5k \\
MNIST &    Full-space &       18M &     22k &     21k &      18M &       10k \\
MNIST &    Full-space &       70M &     43k &     41k &      70M &       20k \\
MNIST &    Full-space &      274M &     84k &     82k &     275M &       40k \\
\midrule
MNIST & Reduced-space &   All NNs &      2k &     805 &      10k &      307k \\
%MNIST & Reduced-space &      167k &      2k &     805 &      10k &      307k \\
%MNIST & Reduced-space &        1M &      2k &     805 &      10k &      307k \\
%MNIST & Reduced-space &        5M &      2k &     805 &      10k &      307k \\
%MNIST & Reduced-space &       18M &      2k &     805 &      10k &      307k \\
%MNIST & Reduced-space &       70M &      2k &     805 &      10k &      307k \\
%MNIST & Reduced-space &      274M &      2k &     805 &      10k &      307k \\
\midrule
SCOPF &    Full-space &       15k &      1k &      1k &      17k &        1k \\
SCOPF &    Full-space &      578k &      4k &      4k &     567k &        2k \\
SCOPF &    Full-space &        4M &     11k &     11k &       4M &        6k \\
SCOPF &    Full-space &       68M &     73k &     73k &      68M &       37k \\
SCOPF &    Full-space &      592M &    305k &    305k &     592M &      153k \\
\midrule
SCOPF & Reduced-space &   All NNs &      1k &      1k &      10k &        8k \\
%SCOPF & Reduced-space &      578k &      1k &      1k &      10k &        8k \\
%SCOPF & Reduced-space &        4M &      1k &      1k &      10k &        8k \\
%SCOPF & Reduced-space &       68M &      1k &      1k &      10k &        8k \\
%SCOPF & Reduced-space &      592M &      1k &      1k &      10k &        8k \\
    \bottomrule
  \end{tabular}
  \label{tab:structure}
\end{table}

\begin{table}[h!]
  \centering
  \rowcolors{1}{gray!10}{white}
  \caption{
  Solve times with different neural networks and formulations}
  \resizebox{\textwidth}{!}{
  \begin{tabular}{ccccccccc}
    \toprule
    Model &   Formulation & Platform & NN param. & Solve time (s) & N. iter. & Time/iter. (s) & Objective \\
\midrule
MNIST &    Full-space &      CPU &      167k &             19 &     290 &           0.07 &       3.3 \\
MNIST &    Full-space &      CPU &        1M &            348 &    1157 &            0.3 &       3.3 \\
MNIST &    Full-space &      CPU &        5M &          11536 &    2110 &              5 &       6.1 \\
MNIST &    Full-space &      CPU &   18M$^*$ &            --  &       --&            --  &       --  \\
\midrule
MNIST & Reduced-space &      CPU &      167k &              3 &      28 &            0.1 &       3.3 \\
MNIST & Reduced-space &      CPU &        1M &              4 &      43 &            0.1 &       3.3 \\
MNIST & Reduced-space &      CPU &        5M &             10 &      77 &            0.1 &       5.9 \\
MNIST & Reduced-space &      CPU &       18M &             63 &     147 &            0.4 &       4.7 \\
MNIST & Reduced-space &      CPU &       70M &             58 &      80 &            0.7 &       2.0 \\
MNIST & Reduced-space &      CPU &      274M &             45 &      31 &              1 &       5.3 \\
\midrule
MNIST & Reduced-space &  CPU+GPU &      167k &              2 &      28 &           0.07 &       3.3 \\
MNIST & Reduced-space &  CPU+GPU &        1M &              4 &      43 &           0.08 &       3.3 \\
MNIST & Reduced-space &  CPU+GPU &        5M &              5 &      67 &           0.08 &       5.9 \\
MNIST & Reduced-space &  CPU+GPU &       18M &             12 &     149 &           0.08 &       4.8 \\
MNIST & Reduced-space &  CPU+GPU &       70M &              8 &      82 &            0.1 &       2.0 \\
MNIST & Reduced-space &  CPU+GPU &      274M &              3 &      28 &            0.1 &       5.3 \\
\midrule
SCOPF &    Full-space &      CPU &       15k &              4 &     675 &           0.01 &     82379 \\
SCOPF &    Full-space &      CPU &      578k &            399 &    1243 &            0.3 &     82379 \\
SCOPF &    Full-space &      CPU &        4M &           8858 &    2505 &              4 &     82379 \\
SCOPF &    Full-space &      CPU &   68M$^*$ &             -- &       --&            --  &       -- \\
\midrule
SCOPF & Reduced-space &      CPU &       15k &              7 &     358 &           0.02 &     82379 \\
SCOPF & Reduced-space &      CPU &      578k &              5 &     147 &           0.03 &     82379 \\
SCOPF & Reduced-space &      CPU &        4M &              3 &      53 &           0.06 &     82379 \\
SCOPF & Reduced-space &      CPU &       68M &             37 &      40 &              1 &     82379 \\
SCOPF & Reduced-space &      CPU &      592M &            144 &      42 &              3 &     82379 \\
\midrule
SCOPF & Reduced-space &  CPU+GPU &       15k &              5 &     298 &           0.02 &     82379 \\
SCOPF & Reduced-space &  CPU+GPU &      578k &              3 &     162 &           0.02 &     82379 \\
SCOPF & Reduced-space &  CPU+GPU &        4M &              1 &      53 &           0.03 &     82379 \\
SCOPF & Reduced-space &  CPU+GPU &       68M &              2 &      40 &           0.04 &     82379 \\
SCOPF & Reduced-space &  CPU+GPU &      592M &              3 &      42 &           0.08 &     82379 \\
    \bottomrule
    \rowcolor{white}\multicolumn{8}{l}{\scriptsize $^*$ Fails to solve within 5 hour time limit}\\[-2pt]
  \end{tabular}
  }
  \label{tab:alltimes}
\end{table}

\subsection{Runtime results}\label{sec:results}

Runtimes for the different formulations with neural network surrogates of increasing size
are given in Table \ref{tab:alltimes}.
For the reduced-space formulation, we also compare CPU-only solves with solves that
leverage a GPU for neural network evaluation (and differentiation).
The results immediately show that the full-space formulation is not scalable to
neural networks with more than a few million trained parameters.
The full-space formulation exceeds the 5-hour time limit on networks above this size.
A breakdown of solve times, given in Table \ref{tab:solvetime-breakdown}, confirms the
bottleneck in this formulation. The full-space formulation spends almost all of its
solve time in the IPOPT algorithm, which we assume is dominated by KKT matrix factorization.
\footnote{This is difficult to confirm directly, but by parsing IPOPT's logs,
we see that, for the SCOPF model with the 4M-parameter neural network, IPOPT reports spending
96\% of its solve time in a category called ``LinearSystemFactorization''.}

By contrast, the reduced-space formulation is capable of solving the optimization problem
with the largest neural network surrogates tested.
While a CPU-only solve takes a relatively long 144 s for the SCOPF problem with a 592M-parameter
neural network embedded, a GPU-accelerated solve of the same problem solves in only three seconds.
In all cases, the solve time with the reduced-space formulation is dominated by
Hessian evaluation, which explains the large speed-ups obtained with the GPU
(9$\times$ for the MNIST problem with a 274M-parameter neural network
and 48$\times$ for the SCOPF problem with a 592M-parameter neural network).
%In fact, the approximate-Hessian solve time with GPU acceleration is still dominated by function
%evaluation, suggesting that further speed-ups would be possible with more powerful hardware.

Finally, we observe that the interior point method requires many more iterations with the
full-space formulation than with the reduced-space formulation. While the theory of
interior point methods' convergence behavior with full and reduced-space formulations
is not well-understood, this behavior is consistent with
lower iteration counts and improved convergence reliability that have been observed
for reduced-space formulations in other contexts \cite{pacaud2022,parker2022,naik2025aggregation}.
In addition to iteration counts, the times-per-iteration are significantly higher
for the full-space formulation, suggesting that its bottlenecks would not be remedied
by using a different optimization algorithm that is able to converge in fewer iterations.

\begin{table}[h!]
  \centering
  \caption{Solve time breakdowns for selected neural networks and formulations}
  \rowcolors{3}{gray!10}{white}
  \resizebox{\textwidth}{!}{
  \begin{tabular}{ccccccccc}
    \toprule
    \multirow{2}{*}{Model} &\multirow{2}{*}{Formulation} &\multirow{2}{*}{NN. param.} &\multirow{2}{*}{Platform} &\multirow{2}{*}{Solve time (s)}& \multicolumn{4}{c}{Percent of solve time (\%)} \\
    \cmidrule(r){6-9}
    & & & & & Function & Jacobian & Hessian & Solver \\
    \midrule
MNIST &    Full-space &        5M &      CPU &          11536 &     0.05 &     0.05 &    0.07 &    99+ \\
MNIST & Reduced-space &      274M &      CPU &             45 &        5 &       13 &      81 &      1 \\
MNIST & Reduced-space &      274M &  CPU+GPU &              3 &        3 &        5 &      72 &     19 \\
\midrule
SCOPF &    Full-space &        4M &      CPU &           8858 &      0.1 &      0.1 &     0.2 &    99+ \\
SCOPF & Reduced-space &      592M &      CPU &            144 &        6 &       14 &      80 &    0.2 \\
SCOPF & Reduced-space &      592M &  CPU+GPU &              3 &        6 &       19 &      68 &      8 \\
    \bottomrule
  \end{tabular}
  }
  \label{tab:solvetime-breakdown}
\end{table}

\section{Conclusion}

This work demonstrates that nonlinear local optimization problems may incorporate neural
networks with hundreds of millions of trained parameters, with minimal overhead,
using a reduced-space formulation that exploits efficient automatic differentiation
and GPU acceleration.
Further research should test this formulation on neural networks with larger input and
output dimensions to measure the point at which CPU-GPU data transfer becomes a bottleneck;
our experiments indicate that, for our test problems, this overhead is small compared to
the effort of evaluating and differentiating the neural network itself.
A disadvantage of our formulation is that it is not suitable
for global optimization as the non-convex neural network constraints
are not represented in a format that can be communicated to any global optimization
solver we are aware of. Interfacing convex under and over-estimators of neural networks
(e.g., CROWN \cite{zhang2018crown}) with global optimization solvers is another
interesting area for future work.
Additionally, relative performance of the full and reduced-space formulations may
change in different applications.
This motivates future research and development to improve the performance
of the full-space formulation, which may be achieved by linear algebra decompositions
that exploit the structure of the neural network's Jacobian in the KKT matrix.

\begin{ack}
  This work was funded by the LDRD program at Los Alamos National Laboratory
  under the Artimis project, the Information Science and Technology Institute,
  and the Center for Nonlinear Studies.
  LA-UR-25-29374.
\end{ack}

\printbibliography

@article{bugosen2024,
  title={Process flowsheet optimization with surrogate and implicit formulations of a {G}ibbs reactor},
  journal={Systems and Control Transactions},
  volume={3},
  pages={113-120},
  year={2024},
  doi={https://doi.org/10.69997/sct.148498},
  author={Sergio I. Bugosen and Carl D. Laird and Robert B. Parker},
}

@article{lopezflores2024,
  author = {López-Flores, Francisco Javier and Ramírez-Márquez, C{\'e}sar and Ponce-Ortega, Jos{\'e} María},
  title = {Process Systems Engineering Tools for Optimization of Trained Machine Learning Models: Comparative and Perspective},
  journal = {Industrial \& Engineering Chemistry Research},
  volume = {63},
  number = {32},
  pages = {13966-13979},
  year = {2024},
  doi = {10.1021/acs.iecr.4c00632},
}

@Article{kilwein2023,
  AUTHOR = {Kilwein, Zachary and Jalving, Jordan and Eydenberg, Michael and Blakely, Logan and Skolfield, Kyle and Laird, Carl and Boukouvala, Fani},
  TITLE = {Optimization with Neural Network Feasibility Surrogates: Formulations and Application to Security-Constrained Optimal Power Flow},
  JOURNAL = {Energies},
  VOLUME = {16},
  YEAR = {2023},
  NUMBER = {16},
  ARTICLE-NUMBER = {5913},
  URL = {https://www.mdpi.com/1996-1073/16/16/5913},
  ISSN = {1996-1073},
  DOI = {10.3390/en16165913}
}

@article{omlt,
author  = {Francesco Ceccon and Jordan Jalving and Joshua Haddad and Alexander Thebelt and Calvin Tsay and Carl D Laird and Ruth Misener},
title   = {{OMLT}: {O}ptimization \& Machine Learning Toolkit},
journal = {Journal of Machine Learning Research},
year    = {2022},
volume  = {23},
number  = {349},
pages   = {1--8},
url     = {http://jmlr.org/papers/v23/22-0277.html}
}

@article{jump1,
author = {Lubin, Miles and Dowson, Oscar and Garcia, Joaquim Dias and Huchette, Joey and Legat, Beno{\^\i}t and Vielma, Juan Pablo},
date = {2023/09/01},
doi = {10.1007/s12532-023-00239-3},
id = {Lubin2023},
isbn = {1867-2957},
journal = {Mathematical Programming Computation},
number = {3},
pages = {581--589},
title = {{JuMP} 1.0: {R}ecent improvements to a modeling language for mathematical optimization},
url = {https://doi.org/10.1007/s12532-023-00239-3},
volume = {15},
year = {2023}
}

@book{griewank2008,
    author = {Griewank, Andreas and Walther, Andrea},
    title = {Evaluating {D}erivatives: {P}rinciples and Techniques of Algorithmic Differentiation},
    year = {2008},
    isbn = {0898716594},
    publisher = {Society for Industrial and Applied Mathematics},
    address = {USA},
    edition = {Second},
}

@article{ipopt,
  title={On the implementation of an interior-point filter line-search algorithm for large-scale nonlinear programming},
  author={W{\"a}chter, Andreas and Biegler, Lorenz T},
  journal={Mathematical programming},
  volume={106},
  number={1},
  pages={25--57},
  year={2006},
  publisher={Springer},
}

@article{paszke2019pytorch,
  title={Pytorch: An imperative style, high-performance deep learning library},
  author={Paszke, Adam and Gross, Sam and Massa, Francisco and Lerer, Adam and Bradbury, James and Chanan, Gregory and Killeen, Trevor and Lin, Zeming and Gimelshein, Natalia and Antiga, Luca and others},
  journal={Advances in neural information processing systems},
  volume={32},
  year={2019}
}

@article{aravena2023,
    author = {Aravena, Ignacio and Molzahn, Daniel K. and Zhang, Shixuan and Petra, Cosmin G. and Curtis, Frank E. and Tu, Shenyinying and W\"{a}chter, Andreas and Wei, Ermin and Wong, Elizabeth and Gholami, Amin and Sun, Kaizhao and Sun, Xu Andy and Elbert, Stephen T. and Holzer, Jesse T. and Veeramany, Arun},
    title = {Recent Developments in Security-Constrained {AC} Optimal Power Flow: {O}verview of Challenge 1 in the {ARPA-E} Grid Optimization Competition},
    journal = {Operations Research},
    volume = {71},
    number = {6},
    pages = {1997-2014},
    year = {2023},
    doi = {10.1287/opre.2022.0315},
}

@manual{powerworldmanual,
  title={{PowerWorld} Simulator Manual},
  organization={PowerWorld Corporation},
  address={Champaign, IL, USA},
  note={https://www.powerworld.com/WebHelp/ Accessed December 10, 2024},
}

@misc{adam,
title={Adam: {A} Method for Stochastic Optimization}, 
author={Diederik P. Kingma and Jimmy Ba},
year={2017},
eprint={1412.6980},
archivePrefix={arXiv},
primaryClass={cs.LG},
url={https://arxiv.org/abs/1412.6980}, 
}

@inproceedings{powermodels,
author = {Carleton Coffrin and Russell Bent and Kaarthik Sundar and Yeesian Ng and Miles Lubin},
title = {PowerModels.jl: An Open-Source Framework for Exploring Power Flow Formulations},
booktitle = {2018 Power Systems Computation Conference (PSCC)},
year = {2018},
month = {June},
pages = {1-8},
doi = {10.23919/PSCC.2018.8442948}
}

@online{powermodelssecurity,
    author = {Carleton Coffrin},
    title = {{PowerModelsSecurityConstrained.jl}},
    year = 2022,
    url = {https://github.com/lanl-ansi/PowerModelsSecurityConstrained.jl},
    note ={Accessed 2024-12-10},
}

@online{hawaii40,
    author = {Adam Birchfield},
    title = {Hawaii Synthetic Grid -- 37 Buses},
    year = 2023,
    url = {https://electricgrids.engr.tamu.edu/hawaii40/},
    note ={Accessed 2024-12-10},
}

@ARTICLE{birchfield2017,
author={Birchfield, Adam B. and Xu, Ti and Gegner, Kathleen M. and Shetye, Komal S. and Overbye, Thomas J.},
journal={IEEE Transactions on Power Systems}, 
title={Grid Structural Characteristics as Validation Criteria for Synthetic Networks}, 
year={2017},
volume={32},
number={4},
pages={3258-3265},
doi={10.1109/TPWRS.2016.2616385},
}

@article{schweidtmann2019deterministic,
  title={Deterministic global optimization with artificial neural networks embedded},
  author={Schweidtmann, Artur M and Mitsos, Alexander},
  journal={Journal of Optimization Theory and Applications},
  volume={180},
  number={3},
  pages={925--948},
  year={2019},
  publisher={Springer}
}

@inproceedings{tjeng2018evaluating,
title={Evaluating Robustness of Neural Networks with Mixed Integer Programming},
author={Vincent Tjeng and Kai Y. Xiao and Russ Tedrake},
booktitle={International Conference on Learning Representations},
year={2019},
url={https://openreview.net/forum?id=HyGIdiRqtm},
}

@inproceedings{bunel2018unified, author = {Bunel, Rudy and Turkaslan, Ilker and Torr, Philip H.S. and Kohli, Pushmeet and Kumar, M. Pawan}, title = {A unified view of piecewise linear neural network verification}, year = {2018}, publisher = {Curran Associates Inc.}, address = {Red Hook, NY, USA}, booktitle = {Proceedings of the 32nd International Conference on Neural Information Processing Systems}, pages = {4795–4804}, numpages = {10}, location = {Montr\'{e}al, Canada}, series = {NIPS'18} }

@book{nocedalwright,
title={Numerical Optimization},
author={Jorge Nocedal and Stephen J. Wright},
publisher={Springer},
year={2006},
}

@article{casas2025comparison,
title = {A comparison of strategies to embed physics-informed neural networks in nonlinear model predictive control formulations solved via direct transcription},
journal = {Computers \& Chemical Engineering},
volume = {198},
pages = {109105},
year = {2025},
issn = {0098-1354},
doi = {https://doi.org/10.1016/j.compchemeng.2025.109105},
author = {Carlos Andrés {Elorza Casas} and Luis A. Ricardez-Sandoval and Joshua L. Pulsipher},
}

@misc{dowson2025moai,
      title={{MathOptAI.jl}: {E}mbed trained machine learning predictors into JuMP models},
      author={Oscar Dowson and Robert B Parker and Russel Bent},
      year={2025},
      eprint={2507.03159},
      archivePrefix={arXiv},
      primaryClass={cs.LG},
      url={https://arxiv.org/abs/2507.03159},
}

@article{lecun1998mnist,
  title={The {MNIST} Database of Handwritten Digits},
  author={LeCun, Yann and Cortes, Corinna and Burges, Christopher J. C.},
  year={1998},
  url={http://yann.lecun.com/exdb/mnist/}
}

@techreport{cain2012history,
  title={History of Optimal Power Flow and Formulations},
  author={Mary B. Cain and Richard P. O'Neill and Anya Castillo},
  year={2012},
  institution={Federal Energy Regulatory Commission},
}

@misc{garcia2025transient,
      title={Transient Stability-Constrained OPF: Neural Network Surrogate Models and Pricing Stability},
      author={Manuel Garcia and Nicole LoGiudice and Robert Parker and Russell Bent},
      year={2025},
      eprint={2502.01844},
      archivePrefix={arXiv},
      primaryClass={math.OC},
      url={https://arxiv.org/abs/2502.01844},
}

@article{ma57,
  author = {Duff, Iain S.},
  title = {{MA57}---a code for the solution of sparse symmetric definite and indefinite systems},
  year = {2004},
  issue_date = {June 2004},
  publisher = {Association for Computing Machinery},
  address = {New York, NY, USA},
  volume = {30},
  number = {2},
  issn = {0098-3500},
  doi = {10.1145/992200.992202},
}

@inproceedings{zhang2018crown,
  author = "Huan Zhang AND Tsui-Wei Weng AND Pin-Yu Chen AND Cho-Jui Hsieh AND Luca Daniel",
  title = "Efficient Neural Network Robustness Certification with General Activation Functions",
  booktitle = "Advances in Neural Information Processing Systems (NuerIPS)",
  year = "2018",
  month = "dec"
}

@inproceedings{applegate2021pdhg,
 author = {Applegate, David and Diaz, Mateo and Hinder, Oliver and Lu, Haihao and Lubin, Miles and O\textquotesingle Donoghue, Brendan and Schudy, Warren},
 booktitle = {Advances in Neural Information Processing Systems},
 editor = {M. Ranzato and A. Beygelzimer and Y. Dauphin and P.S. Liang and J. Wortman Vaughan},
 pages = {20243--20257},
 publisher = {Curran Associates, Inc.},
 title = {Practical Large-Scale Linear Programming using Primal-Dual Hybrid Gradient},
 url = {https://proceedings.neurips.cc/paper_files/paper/2021/file/a8fbbd3b11424ce032ba813493d95ad7-Paper.pdf},
 volume = {34},
 year = {2021}
}

@misc{naik2025aggregation,
      title={Variable aggregation for nonlinear optimization problems},
      author={Sakshi Naik and Lorenz Biegler and Russell Bent and Robert Parker},
      year={2025},
      eprint={2502.13869},
      archivePrefix={arXiv},
      primaryClass={math.OC},
      url={https://arxiv.org/abs/2502.13869},
}

@article{pacaud2022,
  title = {A feasible reduced space method for real-time optimal power flow},
  journal = {Electric Power Systems Research},
  volume = {212},
  pages = {108268},
  year = {2022},
  issn = {0378-7796},
  doi = {https://doi.org/10.1016/j.epsr.2022.108268},
  author = {François Pacaud and Daniel Adrian Maldonado and Sungho Shin and Michel Schanen and Mihai Anitescu},
}

@article{parker2022,
  title = {An implicit function formulation for optimization of discretized index-1 differential algebraic systems},
  journal = {Computers \& Chemical Engineering},
  volume = {168},
  pages = {108042},
  year = {2022},
  issn = {0098-1354},
  doi = {https://doi.org/10.1016/j.compchemeng.2022.108042},
  author = {Robert Parker and Bethany Nicholson and John Siirola and Carl Laird and Lorenz Biegler},
}

@misc{lu2024cupdlp,
      title={cuPDLP.jl: A GPU Implementation of Restarted Primal-Dual Hybrid Gradient for Linear Programming in Julia},
      author={Haihao Lu and Jinwen Yang},
      year={2024},
      eprint={2311.12180},
      archivePrefix={arXiv},
      primaryClass={math.OC},
      url={https://arxiv.org/abs/2311.12180},
}

@manual{cudss,
  title={{NVIDIA cuDSS} (preview): A high-performance {CUDA} Library for Direct Sparse Solvers},
  organization={NVIDIA},
  year={2025},
  url={https://docs.nvidia.com/cuda/cudss/},
}

@INPROCEEDINGS{pacaud2024gpu,
  author={Pacaud, François and Shin, Sungho},
  booktitle={2024 IEEE 63rd Conference on Decision and Control (CDC)},
  title={{GPU}-accelerated dynamic nonlinear optimization with {ExaModels} and {MadNLP}},
  year={2024},
  volume={},
  number={},
  pages={5963-5968},
  doi={10.1109/CDC56724.2024.10886720}}

@article{shin2024accelerating,
title = {Accelerating optimal power flow with GPUs: SIMD abstraction of nonlinear programs and condensed-space interior-point methods},
journal = {Electric Power Systems Research},
volume = {236},
pages = {110651},
year = {2024},
issn = {0378-7796},
doi = {https://doi.org/10.1016/j.epsr.2024.110651},
author = {Sungho Shin and Mihai Anitescu and François Pacaud},
}

\end{document}